\documentclass[10pt,twocolumn,letterpaper]{article}

\usepackage[pagenumbers]{cvpr} 

\usepackage{graphicx}
\usepackage{amsmath}
\usepackage{amssymb}
\usepackage{booktabs}
\usepackage{algorithm}  
\usepackage{algorithmic}

\usepackage[utf8]{inputenc} 
\usepackage[T1]{fontenc}    

\usepackage{multirow}
\usepackage{diagbox}       
\usepackage{url}            

\usepackage{nicefrac}       
\usepackage{microtype}      
\usepackage{xcolor}         
\usepackage{amsmath,amssymb,amsfonts}
\usepackage{threeparttable}
\usepackage{amsthm}
\usepackage{bbding}

\usepackage[pagebackref,breaklinks,colorlinks]{hyperref}

\usepackage[capitalize]{cleveref}
\crefname{section}{Sec.}{Secs.}
\Crefname{section}{Section}{Sections}
\Crefname{table}{Table}{Tables}
\crefname{table}{Tab.}{Tabs.}

\def\confName{CVPR}
\def\confYear{2023}

\begin{document}
\title{Indescribable Multi-modal Spatial Evaluator}

\author{Lingke Kong\\
Manteia Tech\\
{\tt\small kid\_liet@163.com}
\and
X. Sharon Qi\\
University of California, Los Angeles\\
{\tt\small xqi@mednet.ucla.edu}
\and
Qijin Shen\\
Fuzhou University\\
{\tt\small qijinshen@foxmail.com}
\and
Jiacheng Wang\\
Xiamen University\\
{\tt\small jiachengw@stu.xmu.edu}
\and
Jingyi Zhang\\
Xiamen University\\
{\tt\small zhangjingyi1@stu.xmu.edu.cn}
\and
Yanle Hu\footnotemark[1]\\
Mayo Clinic Arizona\\
{\tt\small Hu.Yanle@mayo.edu}
\and
Qichao Zhou\footnotemark[1]\\
Manteia Tech\\
{\tt\small zhouqc@manteiatech.com}}

\maketitle
\renewcommand{\thefootnote}{\fnsymbol{footnote}}
\footnotetext[1]{Corresponding author.}
\renewcommand*{\thefootnote}{\arabic{footnote}}

\begin{abstract}
   Multi-modal image registration spatially aligns two images with different distributions. One of its major challenges is that images acquired from different imaging machines have different imaging distributions, making it difficult to focus only on the spatial aspect of the images and ignore differences in distributions.
   In this study, we developed a self-supervised approach, Indescribable Multi-model Spatial Evaluator (IMSE), to address multi-modal image registration. IMSE creates an accurate multi-modal spatial evaluator to measure spatial differences between two images, and then optimizes registration by minimizing the error predicted of the evaluator. To optimize IMSE performance, we also proposed a new style enhancement method called Shuffle Remap which randomizes the image distribution into multiple segments, and then randomly disorders and remaps these segments, so that the distribution of the original image is changed. Shuffle Remap can help IMSE to predict the difference in spatial location from unseen target distributions. Our results show that IMSE outperformed the existing methods for registration using T1-T2 and CT-MRI datasets. IMSE also can be easily integrated into the traditional registration process, and can provide a convenient way to evaluate and visualize registration results. IMSE also has the potential to be used as a new paradigm for image-to-image translation. Our code is available at \url{https://github.com/Kid-Liet/IMSE}.
\end{abstract}

\section{Introduction}
\label{sec:intro}
The purpose of multi-modal image registration is to align two images with different distributions (Moving ($M$) and Target ($T$) images) by warping the space through the deformation field $\phi$. A major challenge in multi-modal image registration is that images from different modalities may differ in multiple aspects given the fact that images are acquired using different imaging machines, or different acquisition parameters. Due to dramatically different reconstruction and acquisition methods, there is no simple one-to-one mapping between different imaging modalities. From the perspective of measuring similarity, mainstream unsupervised multi-modal registration methods can be divided into two categories: similarity operator based registration and image-to-image translation based registration.

\begin{figure}[t]
	\centerline{\includegraphics[width=0.9\columnwidth]{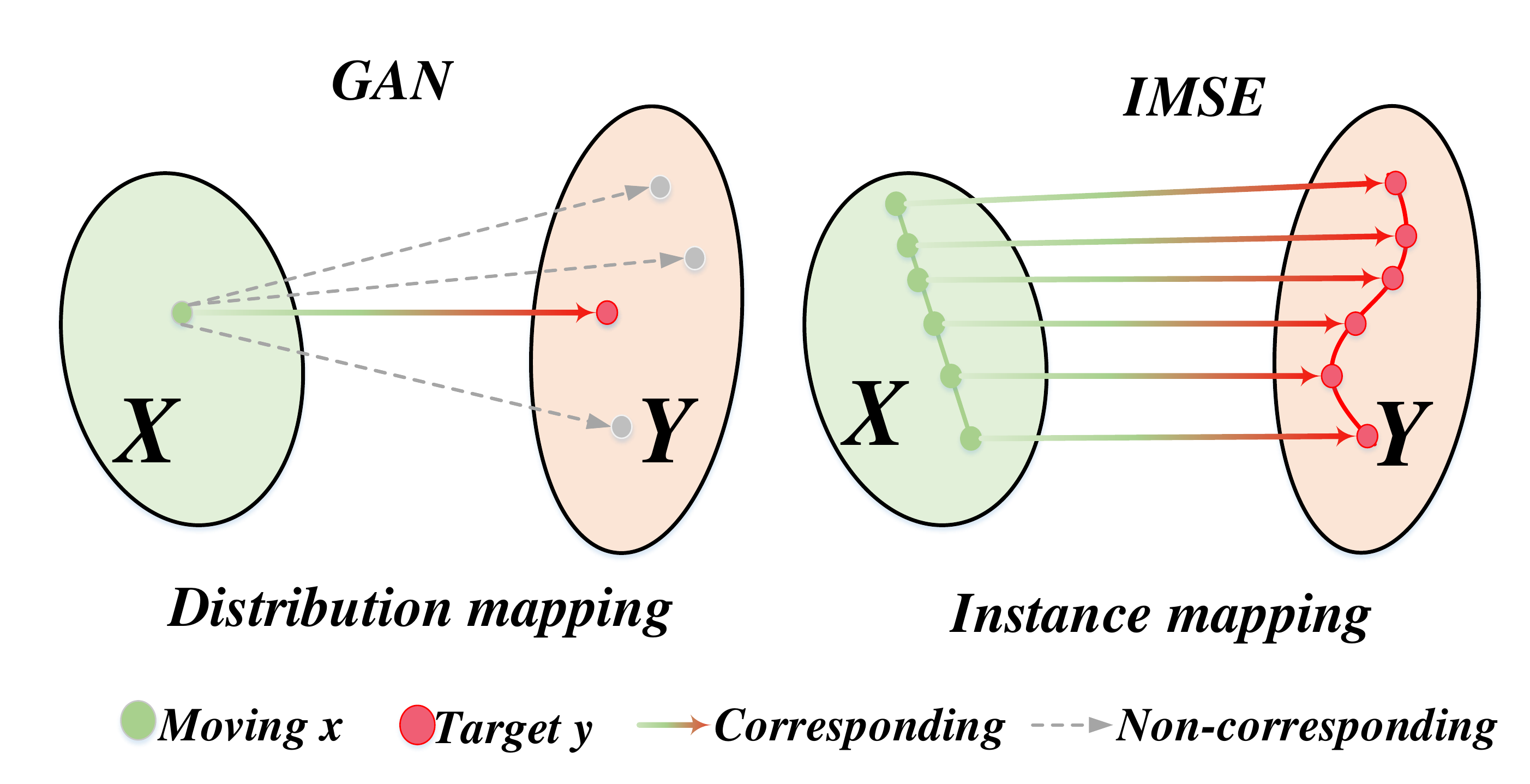}}
	\caption{The GAN based methods can only ensure that the distribution of the $X$ domain is mapped that of the $Y$ domain. Ideally, we want to achieve instance registration in which moving and target images are one-to-one corresponded. }
	\label{fig1}
\end{figure}

\textbf{Similarity operator} based registration uses multi-modal similarity operators as loss functions for registration, for example, normalized cross-correlation (NCC) ~\cite{Cite01,Cite02,Cite03,Cite04}, mutual information (MI)~\cite{Cite05,Cite06,Cite07,Cite08}, and modality-independent neighborhood descriptor (MIND)~\cite{Cite09,Cite10,Cite11,Cite12}. Similarity operators are based on a prior mathematical knowledge. These are carefully designed and improved over time. They can be applied to both traditional registration process (Eq.~\ref{eq1}) and neural network registration (Eq.~\ref{eq2}):
\begin{equation}
\begin{aligned}
\hat{\phi}=\mathop{\arg\min}_{\phi}  \mathcal{L}_{sim}\left(M\left(\phi\right),T\right) .
\end{aligned}
\label{eq1}
\end{equation}
Or
\begin{equation}
\begin{aligned}
\hat{\theta}=\mathop{\arg\min}_{\theta}  \left[\mathbb{E}_{\left(M,T\right)}\left[\mathcal{L}_{sim}\left(M,T, g_{\theta}\left(M,T\right) \right)\right]\right].
\end{aligned}
\label{eq2}
\end{equation}	
Similarity operators have several limitations. \textbf{1)} It is unlikely to design similarity operators that can maintain high accuracy for all data from various imaging modalities. \textbf{2)} It is not possible to estimate the upper limit these operators can achieve and hence it is difficult to find the improvement directions.

\begin{figure}[t]
	\centerline{\includegraphics[width=\columnwidth]{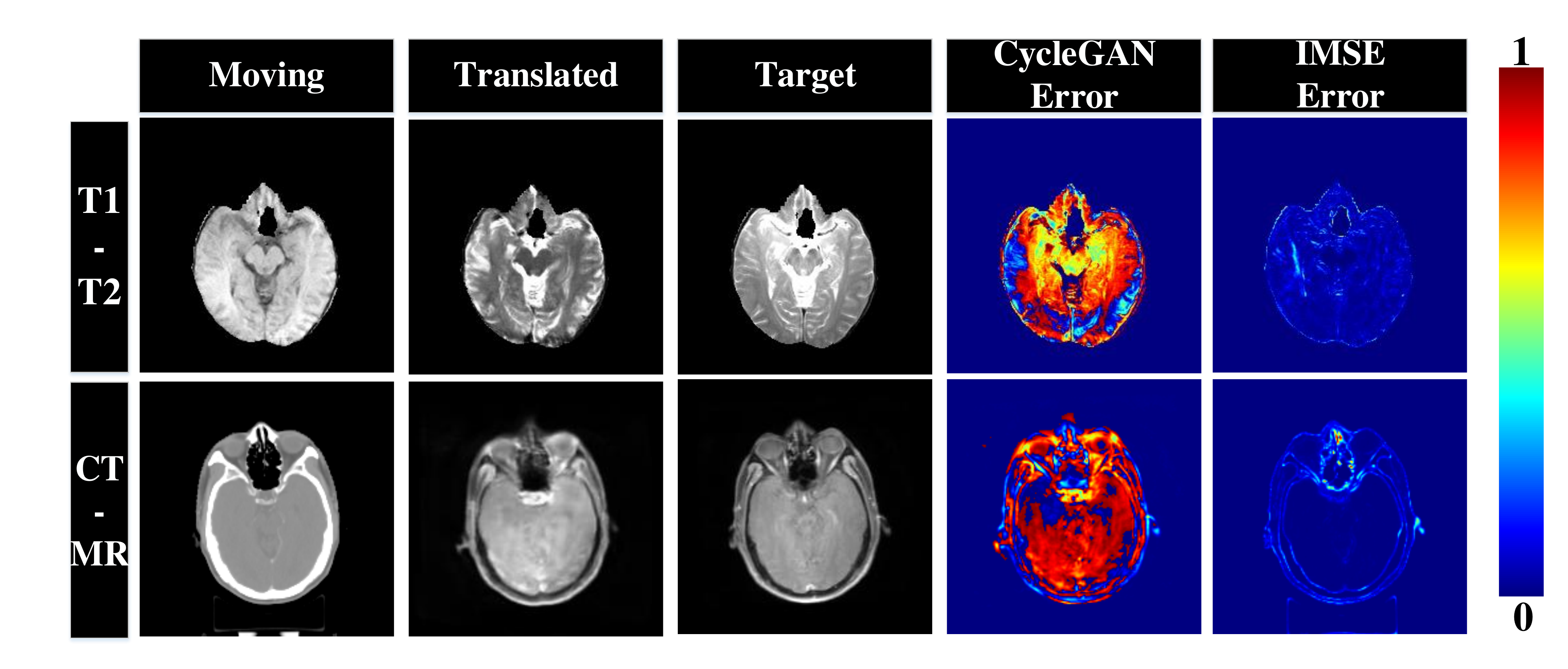}}
	\caption{The distributions of the Moving images translated by CycleGAN still have large residual differences from the Target images. IMSE gives smaller error assessment values in the overlapping regions (converging to blue).}
	\label{fig2}
\end{figure}

\textbf{Image-to-image translation}~\cite{Cite13,Cite14,Cite15,Cite16,Cite17,Cite18,Cite19} based multi-modal registration first translations multi-modal images into single-modal images (Eq~\ref{eq3}) using a generative adversarial network (GAN~\cite{Cite20}), and then use Mean Absolute Error (MAE) or Mean Squared Error (MSE) to evaluate the error at each pixel in space (Eq~\ref{eq4}).
\begin{equation}
\begin{aligned}
\min _{G}\max _{D}\mathcal{L}_{Adv}\left(G,D\right)=\mathbb{E}_{T}\left[log\left(D\left(T\right)\right)\right]+ \\
\mathbb{E}_{M}\left[log \left(1-D\left(G\left(M\right)\right)\right)\right].
\end{aligned}
\label{eq3}
\end{equation}
And
\begin{equation}
\begin{aligned}
\hat{\theta}=\mathop{\arg\min}_{\theta}  \left[\mathbb{E}_{\left(M,T\right)}\left[ \| G\left(M\right),T, g_{\theta}\left(G\left(M\right),T \right)\|_{1} \right] \right].
\end{aligned}
\label{eq4}
\end{equation}
Image-to-image translation based registration cleverly avoids the complex multi-modal problem and reduces the difficulty of registration to a certain extent. However, it has obvious drawbacks. \textbf{1)} The methods based on GAN require training a generator using existing multi-modal data. The trained model will not work if it encounters unseen data, which greatly limits its applicable scenarios. \textbf{2)} More importantly, registration is an instance problem. However, the method based on GAN is to remap the data distribution between different modal. As shown in Figure ~\ref{fig1}, the distribution has bias and variance. We cannot guarantee that the translated image corresponds exactly to the instance target image at the pixel level. For example, Figure~\ref{fig2} shows that there is still residual distribution difference between the target image and translated image. Therefore, even if they are well aligned in space, there is still a large error in the same organ.

To address these challenges, we propose a novel idea based on self-supervision, namely, Indescribable Multi-modal Spatial Evaluator, or IMSE for short. The IMSE approach creates an accurate multi-modal spatial evaluator to metric spatial differences between two images, and then optimizes registration by minimizing
the error predicted by the evaluator. In Figure~\ref{fig2}, we provide a visual demonstration of IMSE in terms of spatial location for T1-T2  and MRI-CT, respectively. Even though distribution differences between the Moving and Target images are still significant, IMSE is low in overlapping regions of the same organs. The main contributions of this study can be summarized as follows: 
\begin{itemize}
    \item 
	We introduce the concepts of relative single-modal and absolute single-modal as an extension of the current definition of single-modality. 
	\item 
	 Based on relative single-modal and absolute single-modal, we propose a self-supervised IMSE method to evaluate spatial differences in multi-modal image registration. The main advantage of IMSE is that it focuses only on differences in spatial location while ignoring differences in multi-modal distribution caused by different image acquisition mechanisms. Our results show that IMSE outperformed the existing metrics for registration using T1-T2 and CT-MRI datasets.
	\item
	 We propose a new style enhancement method named Shuffle Remap.
	 Shuffle Remap can help IMSE to accurate predict the difference in
	 spatial location from an unseen target distribution. As a enhancement method, Shuffle Remap can be impactful in the field of domain generalization.
  	\item
	 We develop some additional functions for IMSE. \textbf{1)} As a measure, IMSE can be integrated into both the registration based on neural network and the traditional registration. \textbf{2)} IMSE can also be used to establish a new image-to-image translation paradigm. \textbf{3)} IMSE can provide provide a convenient way to evaluate and visualize registration results. 
\end{itemize}

\section{Related Work}

\textbf{Multi-modal Similarity Metric:} In order to measure the spatial difference between multi-modal images, several 	classical operators have been proposed. For example, normalized cross-correlation~\cite{Cite01,Cite02,Cite03,Cite04} is used to describe the correlation between two vectors or samples of the same dimension, mutual information~\cite{Cite05,Cite06,Cite07,Cite08} is used to describe the degree of interdependence between variables, and modal independent neighborhood descriptor~\cite{Cite09,Cite10,Cite11,Cite12} is used to describe the local modal characteristics around each voxel. These operators can also be used as loss functions in combination with neural networks~\cite{Cite21,Cite22}. The classical operators are based on mathematical knowledge of researchers. It is difficult to apply to all modal scenes and estimate their upper limits. As GAN~\cite{Cite20} becomes popular in image translation task, researchers have proposed various methods based on GAN to translate multi-modal to single-modal images to facilitate registration. Wei et al.~\cite{Cite23} used CycleGAN~\cite{Cite14} to achieve 2D MR-CT image translation, and converted the 2D slice into 3D volumes through stacking, and finally acted on registration. Qin Chen et al.~\cite{Cite24} proposed an unsupervised multi-modal image-to-image synthesis method by separating the latent shape appearance space and content information space. Kong et al.~\cite{Cite25} proposed a method to add correction to the image translation to improve the quality of the translated image. These GAN based methods remap the data distribution of different modal. There are always biases and variances within the distribution, so registration is an instance problem.

\textbf{Domain Generalization:} The goal of domain generalization (DG) is to generalize to unseen data by training the model on source domain data~\cite{Cite26,Cite27,Cite28,Cite29}. There are methods which aim to learn domain invariant representations by minimizing domain differences between multiple source domains~\cite{Cite30,Cite31,Cite32,Cite33}. In addition, several methods handle DG tasks by modifying the normalization layer, such as instance normalization (IN) and batch normalization (BN)~\cite{Cite34,Cite35,Cite36,Cite37}. In the medical field, Ziqi Zhou et al.~\cite{Cite38} proposed the method of Dual Normalization for segmentation. Wang et al.~\cite{Cite39} used a domain knowledge base to store domain specific prior knowledge, and domain attributes to aggregate the characteristics of different domains. Shuffle Remap proposed by us is a pure style enhancement method, which can be easily combined with other domain generalization methods. It is well suited to situations with large deviation scales, such as CT-MR.


\section{Methodology}

\subsection{Motivation}

In multi-modal registration, the task is greatly simplified if we can isolate and exclude distribution differences, and focus only on spatial differences. In fact, the GAN-based approach intends to eliminate distribution differences by translating the source-domain image to the target-domain image.
However, although the translated image and the target image can be viewed as single-modal relationship, the residual distribution differences may still be significant, making MAE or MSE unsuitable for optimizing registration performance. 
Further, we can define the relationship between such single-modal images as the \textbf{relative single-modal}.

\begin{figure*}[t]
	\centerline{\includegraphics[scale=0.13]{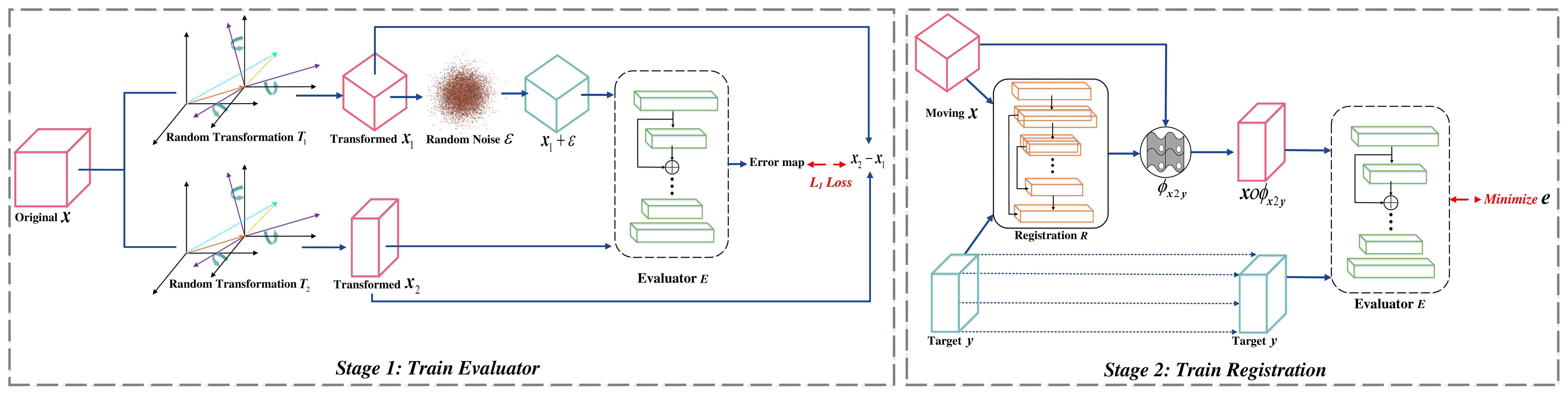}}
	\caption{A general overview of the IMSE process. It is divided into two main parts, the training evaluator and the training registration.}
	\label{fig4}
\end{figure*}

\textbf{Relative single-modal:} For any images $\left\{x_{n}\right\}_{n=1}^{n=N}$, if all satisfy a specific and identical data distribution rule, i.e.  $\left\{x_{n}\right\}_{n=1}^{n=N} \sim \mathcal{D}(\mu,\,\sigma^{2})$, then $\left\{x_{1},...,x_{N}\right\}$ are called relative single-modal data with respect to each other. Relative single-modal is a abstract and extensive concept. For example, if both CT and MR belong to the category of medical images compared with natural images, they can be regarded as relative single-modal images; If T1 and T2  belong to the category of MR compared with CT, they can be regarded as relative single-modal images.

Moreover, we can define the relationship between images without residual distribution difference as \textbf{absolute single-modal}.

\textbf{Absolute single-modal:} For any $(x_{j},x_{k}) \in \left\{x_{n}\right\}_{n=1}^{n=N}$, where $j \neq k$, if $x_{j}$ and $x_{k}$ can be obtained from each other by some particular spatial transformation $\phi$ only, i.e. $x_{j}-x_{k} \circ \phi=0$ or $x_{k}-x_{j} \circ \phi^{-1}=0$. Then $\left\{x_{1},...,x_{N}\right\}$ is called absolute single-modal data. Absolute single-modal belongs to a narrower concept. To some extent, relative single-modal contains absolute single-modal.

The images that belong to absolute single-modal are completely consistent in modalities. We can regard the absolute-modal difference between two images as a spatial error. Therefore, the question becomes how to make the model measure the spatial error from an absolute single-modal perspective.


\subsection{IMSE}
The IMSE method involves training of evaluator and registration separately. In this section, we provide a detailed description of the method. 

\textbf{1.Training evaluator:} In IMSE, we can completely simulate multi-modal registration data and obtain the absolute single-modal label. We first apply two random spatial transformations $T_{1}$ and $T_{2}$ to the original image $x$ to obtain the transformed images $x_{1}$ and $x_{2}$, respectively. The spatial transformation operations include overall rotation, displacement, rescaling and random pixel-wise deformation. Since $x_{1}$ and $x_{2}$ are from the same image $x$, they satisfy the definition of absolute single-modal and differ only in spatial location. Once $x_{1}$ and $x_{2}$ are generated, we subtract the two images to obtain the image of spatial position error, which is used as the label for evaluator training. Next, we add a random noise $\varepsilon$(see Section 3.3 for the specific noise form)  to $x_{1}$ to create distribution differences between $x_{1}$ and $x_{2}$.  
$x_{1} + \varepsilon$ and $x_{2}$ are then stacked according to the number of channels and used as inputs to the evaluator $E$. Then, the predictions of Evaluator and previously created label are used for training (Eq.~\ref{eq5}).
It is worth noting that the distribution of $x_{1} + \varepsilon$ is arbitrary, but $x_{2}$ is unchanged.
Therefore, the trained evaluator take the input $x_{2}$ as the reference image to predict the absolute single-modal error ($x_{2}-x_{1}$).

\begin{equation}
\begin{aligned}
\min _{E}\mathcal{L}_{L_{1}}\left(E\right)=\mathbb{E}_{x_{1},x_{2}}\left[\|E\left(x_{2} , x_{1}+\varepsilon\right),\left(x_{2}-x_{1}\right) \|_{1}\right].
\end{aligned}
\label{eq5}
\end{equation}

\textbf{2.Training registration:} Given the moving image  $x$ and the target image $y$, deformation fields $\phi_{x2y}$ are obtained using a registration network $R$ with $x$ and $y$ as inputs.  The deformation fields are used to obtain the warped image $x\circ\phi_{x2y}$ by warping the moving image $x$. Then, 
the warped image $x\circ\phi_{x2y}$ and the target images $y$ are fed into the evaluator $E$ to get the spatial error  $e$. By minimizing the error $e$, we can optimize the parameters of the registration network $R$ (Eq.~\ref{eq6}). We should also use a regularization constraint based on the deformation fields (Eq.~\ref{eq7}). During the training of the registration network, the parameters of the evaluator remain the same and are not updated. 
\begin{equation}
\begin{aligned}
\min _{R}\mathcal{L}_{sim}\left(R\right)=\mathbb{E}_{x,y}\left[\|E\left(x \circ R\left(x,y \right), y\right) \|_{1}\right].
\end{aligned}
\label{eq6}
\end{equation}
\begin{equation}
\begin{aligned}
\min _{R}\mathcal{L}_{smooth}\left(R\right)=\mathbb{E}_{x,y}\left[\|\nabla R\left(x,y\right)\|^{2}\right].
\end{aligned}
\label{eq7}
\end{equation}

IMSE uses a neural network to evaluate the similarity between multi-modal images. It is not possible to describe its computational process using mathematical formulas like what traditional similarity operators do.  This is analogous to the discriminator in generative adversarial networks. In the training process of the generator, it is impossible to design an analytical operator and use it as a loss function to evaluate the authenticity of the generated images and optimize the generator. To overcome this challenge, the researchers use the classification loss of the discriminator to optimize the generator indirectly. Similarly, IMSE uses the output of the evaluator to update the registration network. Unlike GAN, IMSE has no adversarial process and does not necessarily train both the generator and the discriminator simultaneously as GAN does. IMSE trains the evaluator and registration network separately. Thus, it can save more computational resources.
\subsection{Shuffle Remap}
The next question we need to consider is what factors determine the performance limitations of IMSE. Based on the training process, IMSE belongs to the category of self-supervision. The label used in the training is absolutely accurate. The upper limit of performance that the evaluator can achieve depends mainly on two factors. First, the degree of deformation added to the image needs to sufficiently cover spatial differences presented in the task. Second, the noise added to the image needs to have sufficient diversity and cover the range of distribution differences. 

\begin{algorithm}[t]
	\caption{Pseudocode of Shuffle Remap in PyTorch style.}  
	\label{alg:A}  
	
	\textcolor[rgb]{0,0.5,0.5}{\# X: the input image and the range is [-1,1] }\\
	\textcolor[rgb]{0,0.5,0.5}{\# r\_min: Minimum number of random control points}\\
	\textcolor[rgb]{0,0.5,0.5}{\# r\_max: Maximum number of random control points}\\
	
	\textcolor[rgb]{0,0.5,0.5}{\# number of randomly generated control points}\\
	$ \mathrm{control\_point  \textcolor[rgb]{0.66,0.133,1}{=} random.randint( r\_min, r\_max)} $ 	\\
	\textcolor[rgb]{0,0.5,0.5}{\# normalize to the range of the image distribution}\\
	$ \mathrm{dist  \textcolor[rgb]{0.66,0.133,1}{=} torch.rand(control\_point) \textcolor[rgb]{0.66,0.133,1}{*}(1 \textcolor[rgb]{0.66,0.133,1}{-}( \textcolor[rgb]{0.66,0.133,1}{-}1))  \textcolor[rgb]{0.66,0.133,1}{+} ( \textcolor[rgb]{0.66,0.133,1}{-}1)} $ \\
	\textcolor[rgb]{0,0.5,0.5}{\# sort from small to large}\\
	$ \mathrm{dist  \textcolor[rgb]{0.66,0.133,1}{=} torch.sort(dist)}$\\	
	\textcolor[rgb]{0,0.5,0.5}{\# Add endpoint -1 and 1}\\
	$ \mathrm{dist  \textcolor[rgb]{0.66,0.133,1}{=} torch.cat([torch.tensor([ \textcolor[rgb]{0.66,0.133,1}{-}1]),dist])} $\\
	$ \mathrm{dist  \textcolor[rgb]{0.66,0.133,1}{=} torch.cat([dist,torch.tensor([1])])} $\\
	\textcolor[rgb]{0,0.5,0.5}{\# shuffle the distribution and generate empty new image}\\
	$ \mathrm{shuffle\_remap  \textcolor[rgb]{0.66,0.133,1}{=} torch.randperm(control\_point \textcolor[rgb]{0.66,0.133,1}{+}1)} $\\
	$ \mathrm{new\_X  \textcolor[rgb]{0.66,0.133,1}{=} torch.zeros\_like(x)} $\\
	\textcolor[rgb]{0,0.5,0}{for} i \textcolor[rgb]{0,0.5,0}{in range}(control\_point \textcolor[rgb]{0.66,0.133,1}{+}1):\\
	$\mathrm{ \qquad target\_part  \textcolor[rgb]{0.66,0.133,1}{=} shuffle\_remap[i]} $\\
	$\mathrm{ \qquad min1,max1  \textcolor[rgb]{0.66,0.133,1}{=} dist[i],dist[i \textcolor[rgb]{0.66,0.133,1}{+}1]} $\\
	$\mathrm{ \qquad min2,max2  \textcolor[rgb]{0.66,0.133,1}{=} dist[target\_part],dist[target\_part \textcolor[rgb]{0.66,0.133,1}{+}1]} $\\
	\textcolor[rgb]{0,0.5,0.5}{\qquad \# get the coordinates corresponding to the distribution }\\
	$\mathrm{ \qquad coord \textcolor[rgb]{0.66,0.133,1}{=} torch.where((min1  \textcolor[rgb]{0.66,0.133,1}{<=} X) \textcolor[rgb]{0.66,0.133,1}{\&} (X\textcolor[rgb]{0.66,0.133,1}{<} max1))} $\\
	\textcolor[rgb]{0,0.5,0.5}{\qquad \# Eq.(\textcolor[rgb]{1,0,0}{8})}\\
	$\mathrm{ \qquad new\_X[coord] \textcolor[rgb]{0.66,0.133,1}{=} ((X[coord]\textcolor[rgb]{0.66,0.133,1}{-}min1)/(max1\textcolor[rgb]{0.66,0.133,1}{-}min1))\textcolor[rgb]{0.66,0.133,1}{*}} $\\
	$\mathrm{ \qquad\qquad \qquad\qquad  (max2\textcolor[rgb]{0.66,0.133,1}{-}min2)\textcolor[rgb]{0.66,0.133,1}{+}min2} $\\
	$\mathrm{ \textcolor[rgb]{0,0.5,0.3215}{return}\quad new\_X} $\\
	\label{alog1}
\end{algorithm}
We propose a simple yet effective style enhancement method named Shuffle Remap to ensure sufficient coverage of distribution differences. Specifically, we first normalize the distribution of the original image $X$ to [-1,1], then generate some random control points in between [-1,1] with the endpoints $P_{0}$ and $P_{N}$ fixed at -1 and 1, respectively. These control points randomly divide the image distribution into $N$ segments. Each segment is assigned an index number $n$ in the order from the smallest to the largest. After that, the segments are randomly disordered and remapped according to the disordered order. For example, Eq.~\ref{eq8} demonstrates the remapping from the range $(Pi,Pi+1)$ to the range $(Pj,Pj+1)$ for a given pixel. The complete algorithm flow is given in Algorithm\ref{alog1}.

\begin{equation}
\begin{aligned}
x^{'}=\frac{x-p_{i}}{p_{i+1}-p_{i}}*(p_{j+1}-p_{j})+p_{j}.
\end{aligned}
\label{eq8}
\end{equation}

\begin{figure}[t]
	\centerline{\includegraphics[width=\columnwidth]{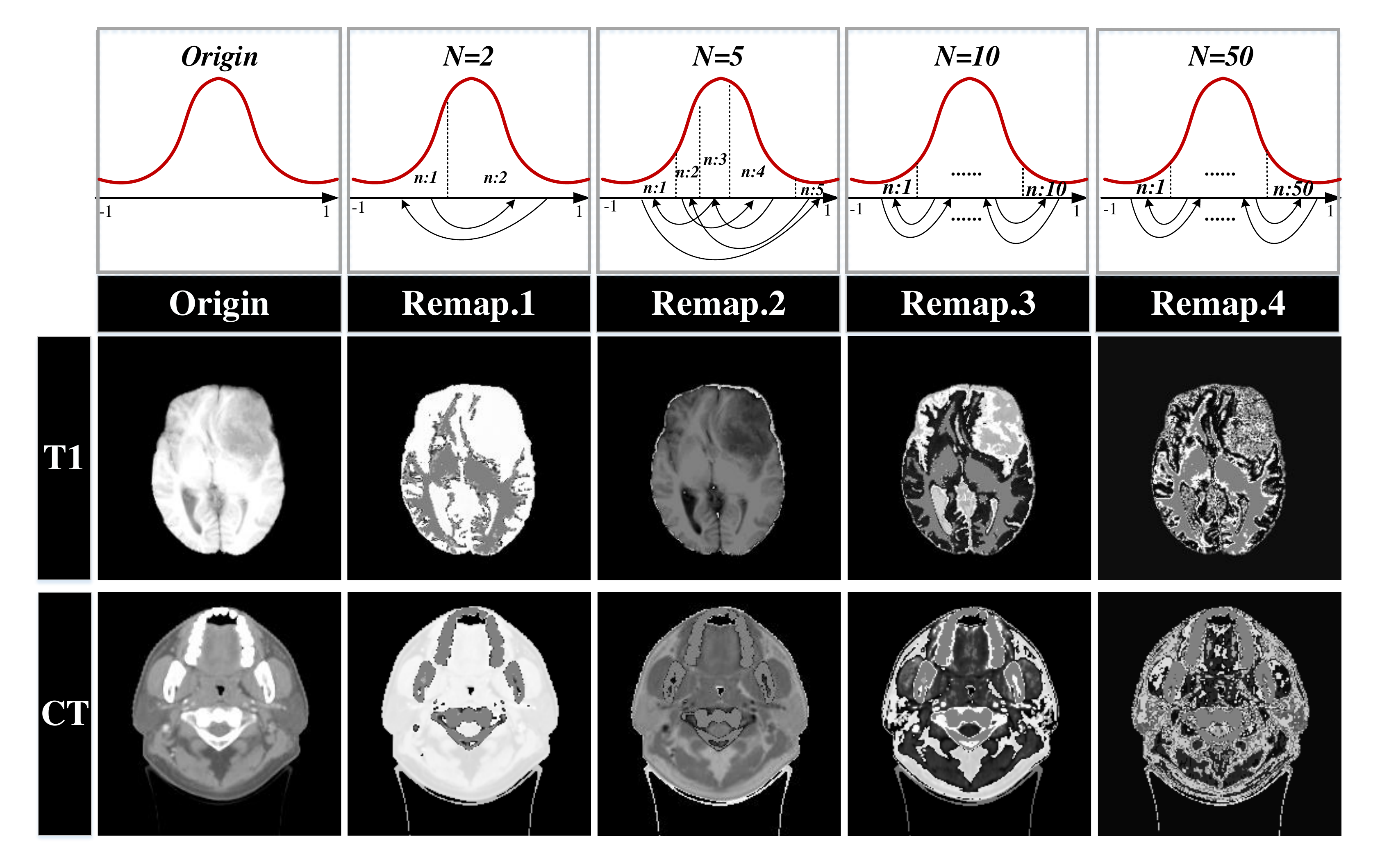}}
	\caption{Examples of T1 and CT after Shuffle Remap.}
	\label{fig5}
\end{figure}
Figure~\ref{fig5} shows the effect of Shuffle Remap. We show the results for different segments. Since the index number and location of control points are random, the remapped images either blur the contrast (Remap.1) or enhance the contrast (Remap.3) between the anatomical structures on the original image. The distribution of remapped images also can be very far away (Remap.2) from the original image distribution. In addition, with the increase of segments, the original image can also be confused (Remap.4). Regardless of Shuffle Remap results, the label used in evaluator training is always unique and accurate, which has the obvious benefit of enhancing the model's knowledge of the same anatomical structures while reducing sensitivity to differences in image distribution. Shuffle Remap is very different from the commonly used histogram shift method. Histogram shift only scales the image distribution without changing the relative relationship of the overall image distribution. Shuffle Remap, however, is a completely random remap of the original image distribution.

It is worth noting that we only provide a style enhancement method, and Shuffle Remap is not irreplaceable in the IMSE architecture. Shuffle Remap is a pure style enhancement method, which can be easily combined with other domain generalization methods.

\section{Experiments}
In this work, we did 4 experiments to evaluate the potential of IMSE. (1) We evaluated the performance of IMSE in multi-modal image registration and compared it with various existing registration methods based on neural networks. (2) We integrated IMSE into the traditional registration procedure. (3) We investigated the feasibility of IMSE as a new image-to-image translation paradigm. (4)We explored the accuracy of using IMSE to assess spatial error.


\subsection{Dataset}
The first dataset is from T1-T2 modal in BraTS2019~\cite{Cite40}, and the second dataset is from clinical CT-MR modal. CT-MR dataset registration accuracy was evaluated based on the parotid gland which was contoured by physicians. Table~\ref{tab1} provides a brief description of the datasets used in this study.
\begin{table}[h]
	\tiny
	\setlength{\tabcolsep}{0.6mm}
	\begin{center}
		\renewcommand\arraystretch{1}
		\begin{tabular}{ccc|cccc|cccccccccc}
			\hline
			\hline
			&&&&\textbf{2D}& & &  &\textbf{3D}\\
			\hline
			\textbf{Source}&\textbf{Modality}&\textbf{Position}&\textbf{Size}&\textbf{Train}&\textbf{Test}&\textbf{Resize}&\textbf{Size}&\textbf{Train}&\textbf{Test}&\textbf{Resize}\\
			\hline
			BraTs 2019&T1-T2&Brain&$240 \times 240$ &18911 & 640  & \XSolidBrush&$48\times128\times 128$ &187 &20&\CheckmarkBold\\
			\hline
			Clinical&CT-MR&Head neck&$192 \times 192$ &3839& 863  &\CheckmarkBold&$48\times128\times 128$ &80 &18 &\CheckmarkBold\\
			\hline
		\end{tabular}
	\end{center}
	\caption{A brief description of the datasets used in the study.}
	\vspace{-3mm}
	\label{tab1}
\end{table}

\begin{table*}[t]
	\begin{minipage}{0.42\linewidth}
		\tiny
		\setlength{\tabcolsep}{0.8mm}
		
		\renewcommand\arraystretch{1}
		\begin{tabular}{c|c|ccc|cccccccccccc}
			\hline
			\hline
			\multirow{3}{*}{Moving$\rightarrow$ Target}&&&2D& & &  3D\\
			\cline{3-8}
			&Methods&Dice $\uparrow$&HD95 $\downarrow$& $\|\nabla \phi\|_{2} \downarrow$& Dice $\uparrow$&HD95 $\downarrow$& $\|\nabla \phi\|_{2} \downarrow$\\
			
			\multirow{9}{*}{T1 $\rightarrow$ T2}& Initial & 0.68 $\pm$ 0.08 &4.17$\pm$ 1.76 &\XSolidBrush &0.81$\pm$ 0.05&4.13$\pm$ 1.53 &\XSolidBrush\\
			\cline{1-8}
			& NCC & 0.75 $\pm$ 0.05  &2.71$\pm$ 1.33 &0.0026 &0.84 $\pm$ 0.03 &3.38 $\pm$ 1.03 &0.0036\\
			& MI &0.82 $\pm$ 0.04 &1.70$\pm$ 1.29&0.0102 & 0.86$\pm$ 0.04 &2.79 $\pm$ 1.44&0.0162	\\
			& MIND &0.83 $\pm$ 0.04&1.66$\pm$ 1.27&\textbf{0.0023}&0.88$\pm$ 0.02&2.70 $\pm$ 1.01&0.0034\\
			& CycleGAN &0.85 $\pm$ 0.03& 1.37$\pm$ 0.95 &0.0061&0.88$\pm$ 0.02&2.94 $\pm$ 0.93&0.010\\
			& RegGAN &0.86 $\pm$ 0.03 & 1.25 $\pm$ 0.90&0.0091 &0.89$\pm$ 0.01&2.85 $\pm$ 0.83&0.0090\\
			& IMSE(BC)&0.83 $\pm$ 0.04 & 1.72 $\pm$ 1.30&0.0125 &0.84$\pm$ 0.04&3.21 $\pm$ 1.33&0.0105\\
			& IMSE(SR)&\textbf{0.89 $\pm$ 0.02} &\textbf{1.06$\pm$ 0.87} &\textbf{0.0023}&\textbf{0.91$\pm$ 0.01} &\textbf{2.36 $\pm$ 0.77} &\textbf{0.0032}\\
			\hline
			\multirow{7}{*}{T2 $\rightarrow$ T1}& NCC & 0.74 $\pm$ 0.04 &2.76$\pm$ 1.35& 0.0026& 0.84$\pm$ 0.03 &3.07 $\pm$ 1.04& 0.0038\\
			& MI &0.79 $\pm$ 0.05 &1.58$\pm$ 1.31&0.0103&0.88$\pm$ 0.04 &2.33 $\pm$ 1.41&0.0166\\
			& MIND & 0.81 $\pm$ 0.03 &2.15$\pm$ 1.28 &0.0023& 0.88$\pm$ 0.02 &2.42 $\pm$ 1.05 &0.0035\\
			& CycleGAN &0.86 $\pm$ 0.04& 1.19$\pm$ 0.94 &0.0056&0.88$\pm$ 0.02& 2.95 $\pm$ 0.95 &0.009\\
			& RegGAN &0.86 $\pm$ 0.03&1.20$\pm$ 0.90 &0.0071 &0.89$\pm$ 0.01&2.71  $\pm$ 0.80&0.0085\\
			& IMSE(BC)&0.80 $\pm$ 0.04 & 1.87 $\pm$ 1.34&0.0122 & 0.85$\pm$ 0.03&3.01 $\pm$ 1.26&0.0097\\
			& IMSE(SR) &\textbf{0.89 $\pm$ 0.02} & \textbf{0.85$\pm$ 0.86} &\textbf{0.0022}&\textbf{0.91$\pm$ 0.01} & \textbf{2.21 $\pm$ 0.75} &\textbf{0.0025}\\
			\hline
		\end{tabular}	
	\end{minipage}
	\hfill
	\begin{minipage}{0.48\linewidth}
		\tiny
		\setlength{\tabcolsep}{0.8mm}
		\renewcommand\arraystretch{1}
		\begin{tabular}{c|c|ccc|cccccccccccc}
			\hline
			\hline
			\multirow{3}{*}{Moving$\rightarrow$ Target}&&&2D& & &  3D\\
			\cline{3-8}
			&Methods&Dice $\uparrow$&HD95 $\downarrow$& $\|\nabla \phi\|_{2} \downarrow$& Dice $\uparrow$&HD95 $\downarrow$& $\|\nabla \phi\|_{2} \downarrow$\\
			
			\multirow{9}{*}{CT $\rightarrow$ MR}& Initial & 0.40 $\pm$ 0.07 &10.05$\pm$ 1.93 &\XSolidBrush &0.48$\pm$ 0.05&5.64$\pm$ 1.41 &\XSolidBrush\\
			\cline{1-8}
			& NCC & 0.49 $\pm$ 0.04  &9.05$\pm$ 1.75 &0.0074 &0.54 $\pm$ 0.03 &5.11 $\pm$ 1.18 &0.017\\
			& MI &0.50 $\pm$ 0.05 &8.90$\pm$ 1.77&0.0075 & 0.55$\pm$ 0.03 &5.14 $\pm$ 1.16&0.019	\\
			& MIND &0.50 $\pm$ 0.04&8.55$\pm$ 1.79&0.0019&0.54$\pm$ 0.02&5.10 $\pm$ 1.07&0.008\\
			& CycleGAN &0.56 $\pm$ 0.04& 8.01$\pm$ 1.69 &0.0022&0.58$\pm$ 0.02&4.62 $\pm$ 0.95&0.010\\
			& RegGAN &0.57 $\pm$ 0.03 & 7.83 $\pm$ 1.63 &0.0020 &0.60$\pm$ 0.01&4.45 $\pm$ 0.90&0.009\\
			& IMSE(BC)&0.50 $\pm$ 0.04 & 8.63 $\pm$ 1.69&0.0072 & 0.55$\pm$ 0.03&5.08 $\pm$ 1.06&0.019\\
			& IMSE(SR) &\textbf{0.61 $\pm$ 0.02} &\textbf{6.92$\pm$ 1.51} &\textbf{0.0017}
			&\textbf{0.62$\pm$ 0.01} &\textbf{4.23 $\pm$ 0.82} &\textbf{0.007}\\
			\hline
			\multirow{7}{*}{MR $\rightarrow$ CT}& NCC & 0.49 $\pm$ 0.05 &9.04$\pm$ 1.75& 0.0070& 0.57$\pm$ 0.03 &5.01 $\pm$ 1.09& 0.013\\
			& MI &0.50 $\pm$ 0.04 &9.18$\pm$ 1.75&0.0076&0.58$\pm$ 0.02 &5.13 $\pm$ 1.15&0.019\\
			& MIND & 0.51 $\pm$ 0.04 &8.86$\pm$ 1.78 &0.0019& 0.56$\pm$ 0.03 &5.05 $\pm$ 1.19 &\textbf{0.007}\\
			& CycleGAN &0.58 $\pm$ 0.04& 7.56$\pm$ 1.66 &\textbf{0.0018}&0.58$\pm$ 0.01& 4.81 $\pm$ 0.89 &0.011\\
			& RegGAN &0.59 $\pm$ 0.02&7.30$\pm$1.61 &0.0021&0.61$\pm$ 0.01&4.62  $\pm$ 0.89&0.009\\
			& IMSE(BC)&0.50 $\pm$ 0.04 & 8.85 $\pm$ 1.71&0.0069 & 0.56$\pm$ 0.04&4.97 $\pm$ 1.16&0.016\\
			& IMSE(SR) &\textbf{0.60 $\pm$ 0.01} & \textbf{7.26$\pm$ 1.58} &0.0020&\textbf{0.62$\pm$ 0.01} & \textbf{4.45 $\pm$ 0.86} &0.008\\
			\hline
		\end{tabular}
	\end{minipage}
	\caption{Registration results of various methods based on the T1-T2 and CT-MR dataset. Initial indicates the results before registration. The source data used to train the IMSE were T1 and CT.}
	\label{tab2}
\end{table*}

\subsection{IMSE for Registration Based on Neural Network}
In this subsection, we compared various registration methods based on neural networks. Baseline uses traditional similarity operators as loss functions to update the registration network, including \textbf{NCC}~\cite{Cite03}, \textbf{MI}~\cite{Cite05}, and \textbf{MIND}~\cite{Cite09}. There are also GAN based methods which do translation first and then use MAE as the loss function of the registered network, including \textbf{CycleGAN}~\cite{Cite14}, \textbf{RegGAN}~\cite{Cite15}. We also compared histogram shift using the Bézier curve~\cite{Cite41} (\textbf{IMSE (BC)}) and Shuffle Remap (\textbf{IMSE (SR)}) for style enhancement. The random range of N in Shuffle Remap is [2, 50].

\begin{figure}[t]
	\centerline{\includegraphics[width=\columnwidth]{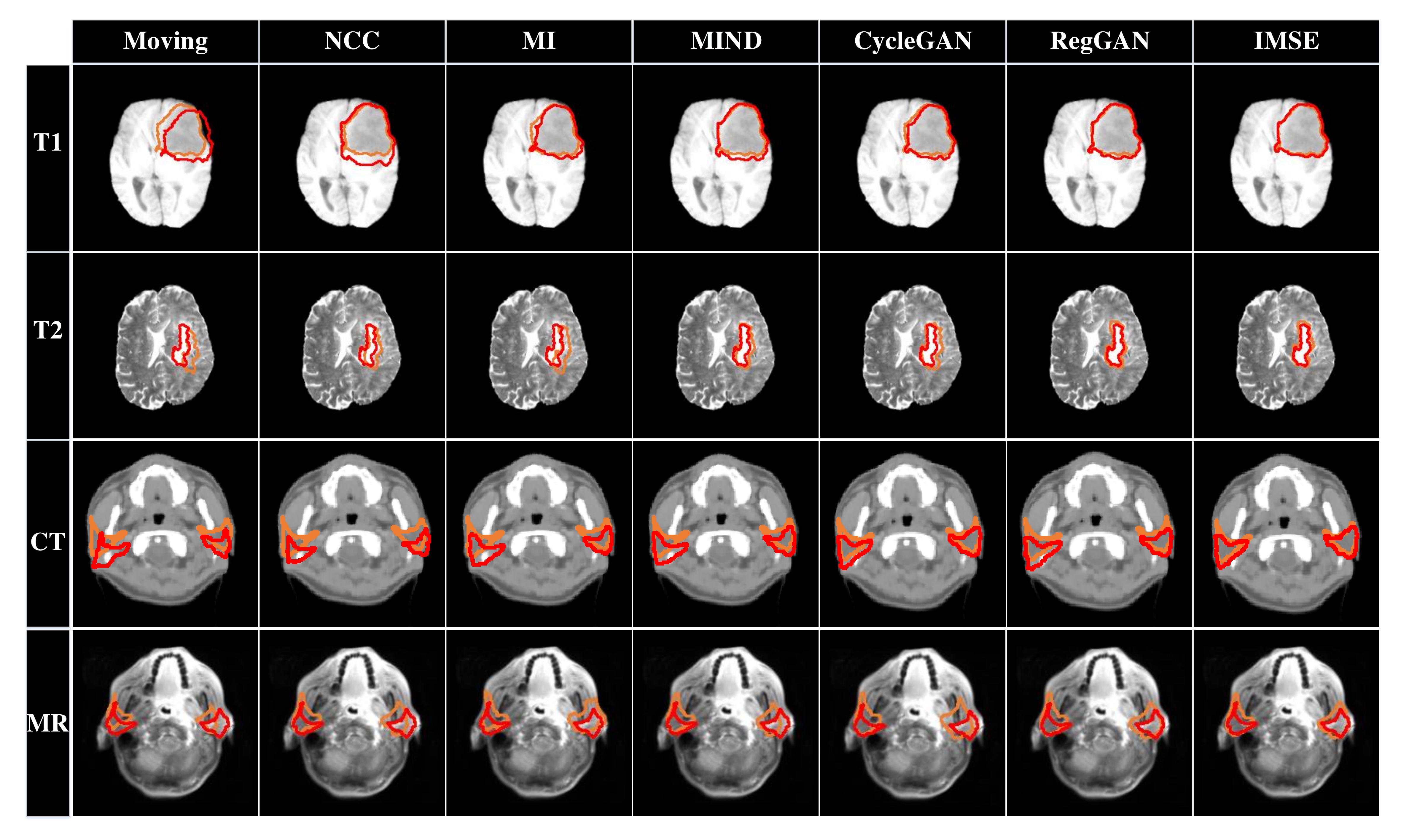}}
	\caption{Registration results for various registration methods. Four rows correspond to T1$\rightarrow$T2, T2$\rightarrow$T1, CT$\rightarrow$MR and MR$\rightarrow$CT registrations, respectively. Orange contours are based on T1 and CT images. Red contours are based on T2 and MR images. The first column (Moving) shows the contour difference without registration.}
	\label{fig6}
\end{figure}

For a fair comparison, all methods used a unified registration network model–VoxelMorph~\cite{Cite21}. Please note that in the two datasets, the source data used to train the estimator were T1 and CT, respectively. The network structure adopted by the evaluator is ResNet~\cite{Cite42}, which is consistent with the generator used in CycleGAN and RegGAN. We added random affine and non-affine transformations to the moving and target images during training and testing, including angular rotations of [-3,3], displacement of [-8\%,+ 8\%], scaling of [8\%,+8\%]. The non-affine transformation was generated by spatially transforming the moving and target images using elastic transformations followed by Gaussian smoothing. The degree of deformation was 80 and the radius of Gaussian smoothing was 12.

Registration results of various methods based on T1-T2 and MR-CT datasets were summarized in Table~\ref{tab2}. We performed both forward and reverse registration for each dataset. Registration performance was measured using Dice, Hausdorff distances, and the smoothness which was defined by the average gradient of the deformation field (in the case of 2D, we only counted the slices that contained contours). For T1-T2 and MR-CT datasets, IMSE achieved the best results based on all metrics, both in 2D and 3D. 
Because the registration network used by all methods was the same, the influence of the model was excluded. IMSE had both high registration accuracy and a smoother deformation field, which in general is difficult to achieve simultaneously. It was likely due to the accurate estimate of spatial errors with the adoption of the estimator, which allowed the registration model to achieve a better alignment at a small deformation cost. In addition, our results show that when the Bézier curve was used for data enhancement, it has no performance advantage compared with other methods. But the combination of IMSE and Shuffle Remap provided the best performance among all registration methods. We want to further explore how different segments N of shuffle remap affect the registration results. As shown in Table~\ref{tab3}, we can first see that with the increase of N, the performance will improve.  In addition,
 for T1-T2, there is no significant difference between N of 30 and 50. For MR-CT, a larger N obviously brings better scores since MR-CT shows larger distribution differences than T1-T2, which requires more complex style enhancement.

\begin{table}[t]
	\begin{minipage}{0.42\linewidth}
	\tiny
	\setlength{\tabcolsep}{0.8mm}
	\renewcommand\arraystretch{1}
	\begin{center}
		\renewcommand\arraystretch{1}
		\begin{tabular}{c|cccccccc}
			\hline
		
			\ T2 $\rightarrow$ T1 &N=2&N=[2,10] & N=[2,30] &N=[2,50]\\

			\hline
			Dice& 0.85& 0.88 &\textbf{0.89} &\textbf{0.89}\\
			HD95& 1.21 &0.93 &0.88 &\textbf{0.85}\\
			$\|\nabla \phi\|_{2}$&0.0028 &0.0024&\textbf{0.0018} &0.0022\\

			\hline
		\end{tabular}
	\end{center}
	\end{minipage}
	\hfill
	\begin{minipage}{0.49\linewidth}
		\tiny
		\setlength{\tabcolsep}{0.8mm}
		\renewcommand\arraystretch{1}
		\begin{center}
			\renewcommand\arraystretch{1}
			\begin{tabular}{c|cccccccc}
				\hline
				
				\ MR $\rightarrow$ CT &N=2&N=[2,10] & N=[2,30] &N=[2,50]\\

				\hline
				Dice& 0.51 & 0.55 & 0.57 & \textbf{0.60} \\
				HD95& 8.83 & 8.21 & 7.49 & \textbf{7.26}\\
				$\|\nabla \phi\|_{2}$&0.020 &0.0029 &0.0025&\textbf{0.0020}\\
				
				\hline
			\end{tabular}
		\end{center}
	\end{minipage}

	\caption{In 2D case, shuffle remap adopts different parameter N}
	\vspace{-4mm}
	\label{tab3}
\end{table}

\subsection{IMSE for Traditional Registration}

Compared to the existing deep learning registration methods which directly provide deformation fields, IMSE essentially evaluates the similarity between multi-modal images through neural networks. Then, the neural network can accurately achieve backward propagation. Therefore, IMSE can be readily integrated into the traditional registration process, such as replacing the $L_{sim}$ function in Eq~\ref{eq1} with a trained evaluator. By first initializing a deformation field of 0 and then optimizing it through similarity loss (Similarity operator or IMSE) and regularization loss(Eq~\ref{eq7}). We set the deformation field size to [64,128,128], the learning rate to 1, the number of iterations to 200, and the optimizer to $adam$. In this section, we not only evaluate the traditional multi-modal registration, but also compare IMSE with single-modal registration using MAE as an optimization measure. 
Single-modal registration is evaluated using T1$\rightarrow$T1 and CT$\rightarrow$CT data.
Since the single-modal images were from different patients with spatial differences significantly larger than those from the same patient, we screened 5 patients whose spatial location differences were relatively small to mimic the scenario of registering images from the same patient. The results are shown in Table~\ref{tab4}. As the metric for registration optimization, IMSE still achieved the best performance in multi-modal conditions. We focus on the results of single-modal registration. IMSE performed much better than MAE in all aspects. This is because even in single-modal datasets, there are still residual distribution differences. T1-T1 or CT-CT should be categorized as relatively single-modal data, especially when images from different patients are registered to each other. MAE cannot ignore residual distribution differences whereas IMSE can. 

\begin{table}[t]

	\setlength{\tabcolsep}{0.5mm}
	\begin{center}
		\renewcommand\arraystretch{1.3}
		\begin{tabular}{c|c|ccccccc}
			\hline
			\hline
			\multirow{1}{*}{Moving$\rightarrow$Target}&Methods&Dice $\uparrow$&HD95 $\downarrow$& $\|\nabla \phi\|_{2} \downarrow$\\
			\hline
			\multirow{4}{*}{T2 $\rightarrow$ T1}& NCC &0.84 $\pm$ 0.03&3.40$\pm$ 1.14 &0.016 \\
            & MI &0.85 $\pm$ 0.03 &3.26$\pm$ 1.22 &0.06\\
            & MIND &0.89 $\pm$ 0.02 &3.01$\pm$ 1.16 &\textbf{0.002}\\
			& IMSE &\textbf{0.91$\pm$ 0.01} &\textbf{2.62$\pm$ 0.81} &\textbf{0.002}\\
			\hline
            \multirow{4}{*}{MR $\rightarrow$ CT}& NCC &0.54 $\pm$ 0.03&5.18$\pm$ 1.20 &0.011 \\
            & MI &0.55 $\pm$ 0.02 &5.29$\pm$ 1.18 &0.05\\
            & MIND &0.55 $\pm$ 0.02 &4.93$\pm$ 1.24 &0.008\\
			& IMSE &\textbf{0.61$\pm$ 0.01} &\textbf{4.37$\pm$ 0.81} &\textbf{0.007}\\
			\hline
   
			\multirow{2}{*}{T1 $\rightarrow$ T1}
			& MAE &0.64 $\pm$ 0.03 &8.38$\pm$ 1.43 &0.018\\
			& IMSE &\textbf{0.67 $\pm$ 0.02} &\textbf{7.4$\pm$ 1.22} &\textbf{0.005}\\

			\hline
			\multirow{2}{*}{CT $\rightarrow$ CT}
			& MAE &0.56 $\pm$ 0.02 &4.92$\pm$ 1.07 &0.012\\
			& IMSE &\textbf{0.59 $\pm$ 0.02} &\textbf{4.80$\pm$0.87} &\textbf{0.005}\\
			
			\hline
		\end{tabular}
	\end{center}
	\caption{Comparison of registration results for traditional algorithms.}
	\vspace{-3mm}
	\label{tab4}
\end{table}

\subsection{IMSE for Image-to-image Translation}
\begin{figure}[t]
	\centerline{\includegraphics[width=\columnwidth]{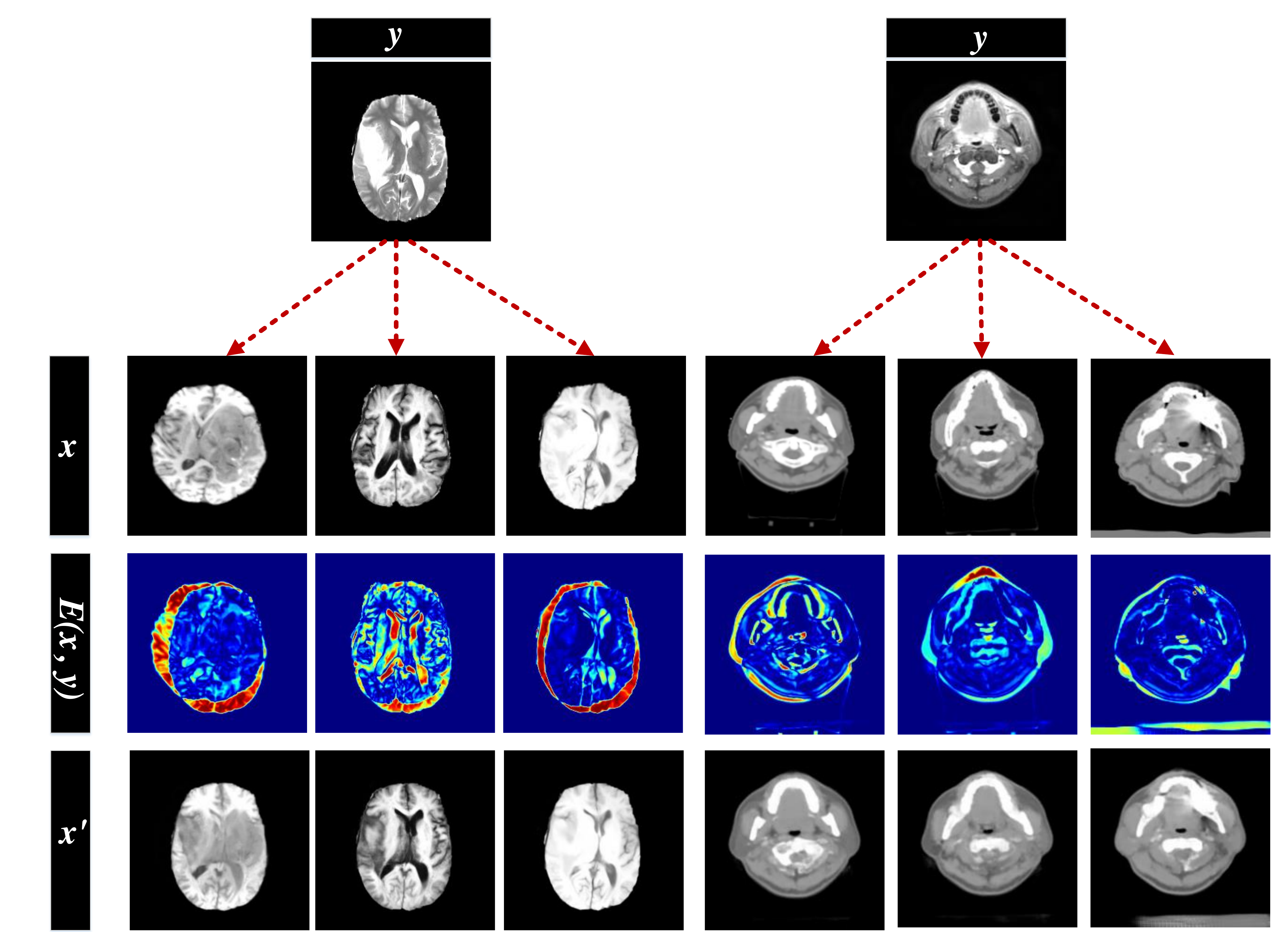}}
	\caption{ IMSE is used for image to image translation. Where, $y$ is the source image, $x$ is different reference images, $E(x,y)$ is the prediction result of IMSE and $x^{'}=x-E(x,y)$.}
	\label{fig7}
\end{figure}

To explain how IMSE may enable a new paradigm of image-to-image translation, we use multi-modal images $x$ and $y$ as an example where $x$ is the reference image and $y$ is the image awaiting to be translated. 
Input $x$ and $y$ into IMSE, and IMSE will use x as the reference image to predict the absolute single-modal error:$E(x,y)\approx x- x^{'}$. Where, the distribution of $x$ and $x^{'}$ is consistent. By subtracting $E(x,y)$ from the reference image $x$, we can get the translated image $x^{'}$ , i.e., $ x^{'}\approx x-  E(x,y)$. 
In the new paradigm, image-to-image translation is achieved by reverse inference based on the prediction of the evaluator.
In Figure~\ref{fig7}, we show a few examples of using IMSE to perform image-to-image translation. $ x^{'}$ is the modal translated image of $y$ as it has the spatial characteristics of $y$ but the modal characteristics of $x$. For example, the dental artifact in the reference CT image in Figure~\ref{fig7} remains in the translated image. This also verifies that IMSE is an instance mapping method based on absolute single-modal.

\begin{table}[t]

	\setlength{\tabcolsep}{0.5mm}
	\begin{center}
		\renewcommand\arraystretch{1.3}
		\begin{tabular}{c|c|ccccccc}
			\hline
			\hline
			\multirow{1}{*}{Source$\rightarrow$Target}&Methods&NMAE $\downarrow$&PSNR $\uparrow$& SSIM $\uparrow$\\
			\hline
			\multirow{3}{*}{T2 $\rightarrow$ T1}& CycleGAN &0.088 &24.1 &0.89 \\
                                                    & RegGAN &0.071  &25.5 &0.90\\
                                        			& IMSE &\textbf{0.029} &\textbf{91.6} &\textbf{0.96}\\
                                        			\hline
                                                    \multirow{3}{*}{MR $\rightarrow$ CT}& CycleGAN &0.049 &22.9 &0.88 \\
                                                    & RegGAN &0.041 &24.1 &0.89\\
                                        			& IMSE &\textbf{0.022} &\textbf{41.3} &\textbf{0.93}\\
			\hline
		\end{tabular}
	\end{center}
	\caption{The results of image-to-image translation.}
	\vspace{-3mm}
	\label{tab5}
\end{table}

Table~\ref{tab5} shows the comparison results between IMSE and baseline methods.  
Spatial transformations were added to $x$ and $y$ in IMSE to prevent alignment between input images. Despite this, IMSE still produces superior results compared to other methods due to the use of an additional target reference image. Thus, the comparison between IMSE and GAN-based methods seems unjust. We aim to investigate the contrasting use scenarios of these two methods, further.

IMSE based image-to-image translation differs from GAN based translation in two aspects. \textbf{1)} The result of image-to-image translation based on GAN is unique whereas IMSE based depends on the characteristics of the reference image. \textbf{2)} GAN based image-to-image translation requires two modal data whereas IMSE based translation only uses one modal data for training. IMSE is very promising for medical image-to-image translation. It requires less data for training and therefore can reduce expensive data costs. It is better suited to the complexity and diversity of medical scenes. Also, medical image translation requires high accuracy. It is always beneficial to ensure that the characteristics of the translated image come from the desired reference image.

\subsection{ IMSE for Spatial Error}
\begin{figure}[t]
	\centerline{\includegraphics[width=\columnwidth]{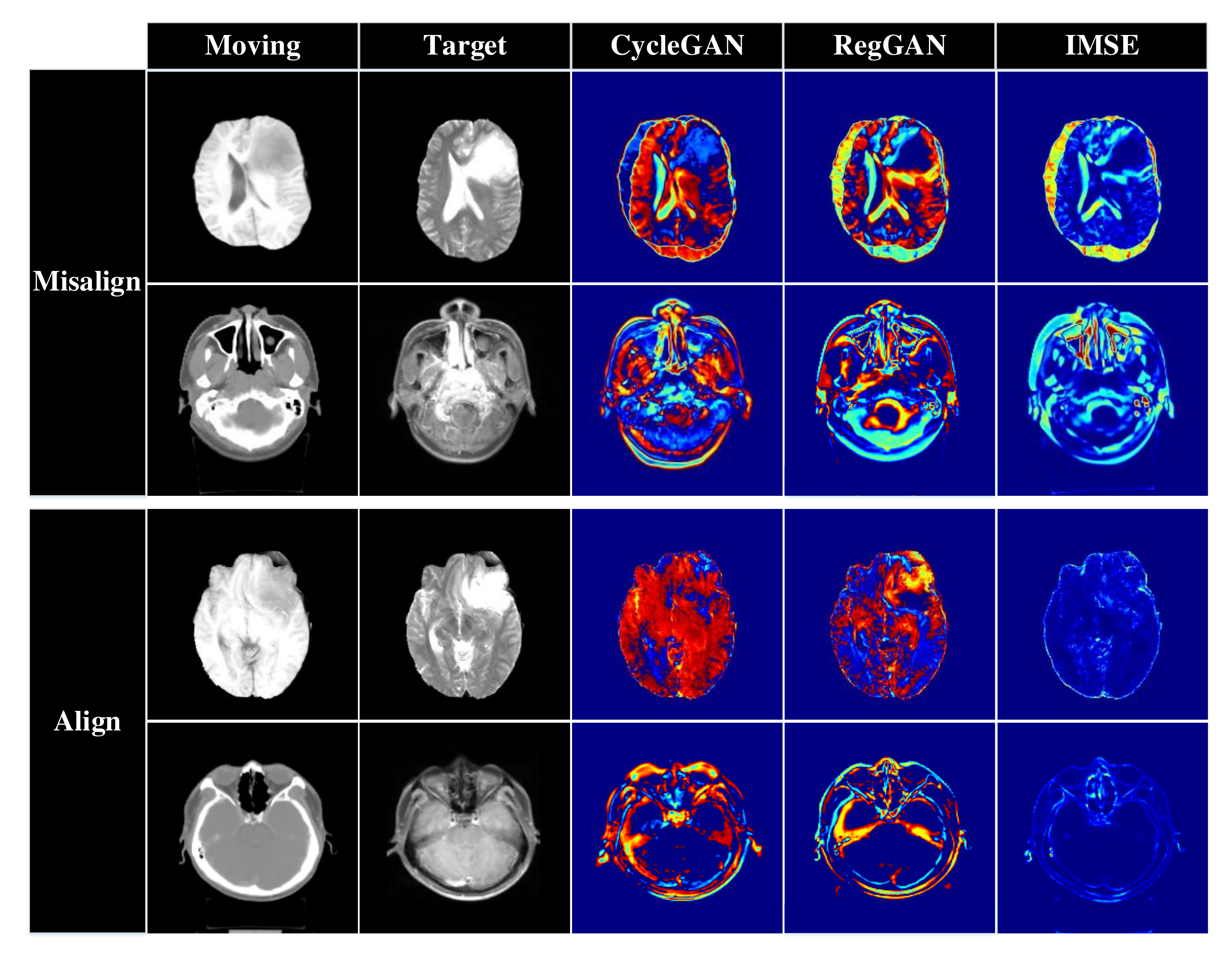}}
	\caption{ Demonstration of  errors estimated by the CycleGAN, RegGAN and IMSE methods in both misaligned and aligned cases. }
	\label{fig8}
\end{figure}

Without labels, accurate evaluation of registration results is always challenging. IMSE has great potential in offering an objective metric to accurately evaluate registration performance. Since the output of IMSE has the same size as the input image, it can provide pixel-wise registration error estimation. In Figure~\ref{fig8}, we demonstrate spatial errors estimated by the CycleGAN, RegGAN and IMSE methods in both misaligned and aligned cases. CycleGAN and RegGAN translated the moving image first and then calculate MAE between the translated image and the target image. 

The estimated spatial errors in the CycleGAN and RegGAN methods could not exclude the effect of residual distribution  between the translated and target images. The errors remained significant even if images were well aligned. As a comparison, IMSE reported large spatial errors in misaligned regions but small spatial errors in aligned regions. 

\begin{figure}[t]
	\centerline{\includegraphics[width=\columnwidth]{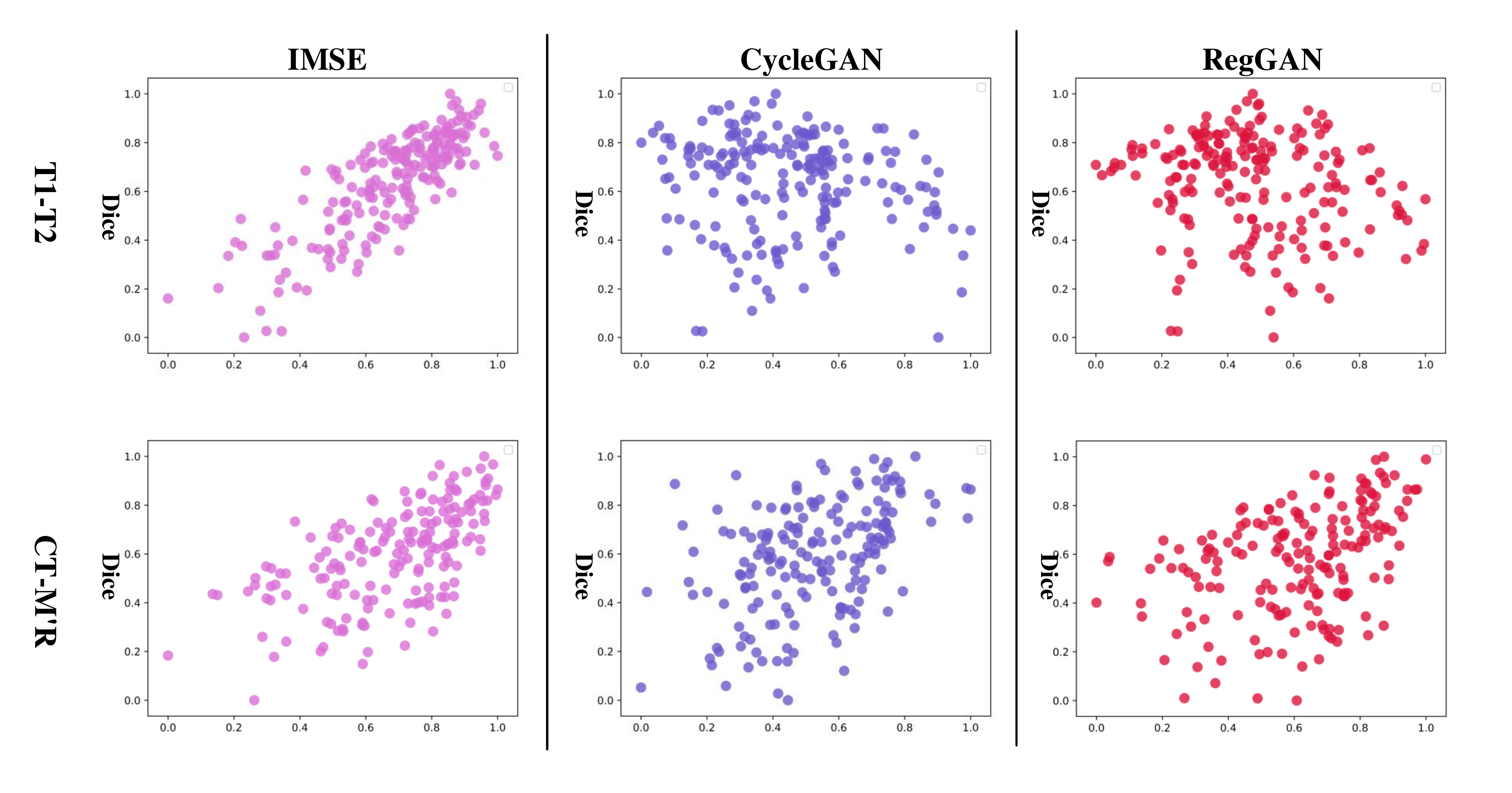}}
	\caption{ Correlation of estimated spatial errors with Dice for the IMSE, CycleGAN, RegGAN methods.}
	\label{fig9}
\end{figure}
To quantitatively demonstrate the potential of IMSE for registration performance evaluation, we explored the correlation between IMSE output and Dice. Based on the joint region of Moving and Target masks, we calculated the mean value of IMSE outputs. Then we subtracted the mean absolute value from 1 and performed a normalization to get the value which indicated registration performance for IMSE. Similarly, we calculated values indicating registration performance for CycleGAN and RegGAN. The tests were performed in 3D due to limited test data. To increase the number of test data, we simulated more test data through many random transformations of the image. Figure~\ref{fig9} clearly demonstrates the positive correlation between IMSE and Dice, which confirms the potential of using IMSE to accurately evaluate spatial errors. As a comparison, CycleGAN and RegGAN do not have obvious correlation with Dice, especially on the T1-T2 dataset.

\section{Discussion}
In this study, we propose a new approach IMSE for multi-modal image registration. IMSE is simple yet powerful. It can be used as a training registration model. As a metric, IMSE can be used to evaluate the registration results or be combined with traditional registration process. It can also be used to perform image-to-image translation. IMSE uses neural networks instead of similarity operators as loss functions to achieve better results. It also stimulates thoughts of using neural network as indirect constraints to solve challenging problems, instead of committing significant time and efforts searching for operators or loss functions for better performance. Shuffle Remap is an essential component of IMSE. It greatly reduces the amount of data required for model training. All trained models in the current study used only one modal of data and the evaluator has the capability to evaluate and translation unseen data as well. Even though our study focused on medical images, the principle of IMSE should apply to natural images as well. 
In the future, we will continue investigating IMSE from three aspects. \textbf{1)} We have demonstrated the correlation between IMSE and Dice. But the correlation is not as strong as we prefer. We will explore various ways of calculating spatial errors to see if it is possible to improve correlation. \textbf{2)} The evaluator is based on neural networks. If it is integrated into the traditional algorithm, backward propagation requires more time. We will explore whether the evaluator can achieve similar results using a simpler architecture. \textbf{3)} We will explore other variants of Shuffle Remap and prove that Shuffle Remap as a method of style enhancement can be more impactful in domain generalization.
{\small
\bibliographystyle{ieee_fullname}
\bibliography{egbib}
}

\end{document}